# Articulation and Clarification of the Dendritic Cell Algorithm

Julie Greensmith[1], Uwe Aickelin[1], and Jamie Twycross[1]

CS&IT, University of Nottingham, UK, NG8 1BB.
jqg, uxa, jpt@cs.nott.ac.uk

**Abstract.** The Dendritic Cell algorithm (DCA) is inspired by recent work in innate immunity. In this paper a formal description of the DCA is given. The DCA is described in detail, and its use as an anomaly detector is illustrated within the context of computer security. A port scan detection task is performed to substantiate the influence of signal selection on the behaviour of the algorithm. Experimental results provide a comparison of differing input signal mappings.

## 1 Introduction

Artificial immune systems (AIS) are a collection of algorithms developed from models or abstractions of the function of the cells of the human immune system. The first, and arguably the most obvious, application for AIS is in the protection of computers and networks, through virus and intrusion detection[2]. In this paper we present an AIS approach to intrusion detection based on the Danger Theory, through the development of an algorithm based on the behaviour of Dendritic Cells (DCs). DCs have the power to suppress or activate the immune system through the correlation of signals from an environment, combined with location markers in the form of antigen. A DCs function is to instruct the immune system to act when the body is under attack, policing the tissue for potential sources of damage. DCs are natural anomaly detectors, the sentinel cells of the immune system, and therefore the development of a DC based algorithm was only a matter of time. The Dendritic Cell Algorithm (DCA) was introduced in 2005 and has demonstrated potential as a classifier for a static machine learning data set[4] and anomaly detector for real-time port scan detection[5]. The DCA differs from other AIS algorithm for the following reasons:

- multiple signals are combined and are a representation of environment or context information
- signals are combined with antigen in a temporal and distributed manner
- pattern matching is not used to perform detection, unlike negative selection[6]
- cells of the innate immune system are used as inspiration, not the adaptive immune cells and unlike clonal selection, no dynamic learning is attempted

The aim of this paper is to demonstrate the anomaly detection capabilities of the DCA and to clarify which features of the algorithm facilitate detection.

## 2  Dendritic Cells in vivo

The DCA is based on the function of dendritic cells whose primary role is as an antigen presenting cell. DCs behave very differently to the cells of the adaptive immune system. Before describing the function of the algorithm we give a general overview of DC biology, introducing different cells, organs and their behaviour. More information on natural DCs can be found in [9].

In vivo, DCs can perform a number of different functions, determined by their state of maturation. Modulation between these states is facilitated by the detection of signals within the tissue - namely danger signals, PAMPs (pathogenic associated molecular patterns), apoptotic signals (safe signals) and inflammatory cytokines which are described below. The maturation state of a DC is determined by the relative concentrations of these four types of signal. The state of maturity of a DC influences the response by T-cells, to either an immunogenic or tolerogenic state, for a specific antigen. Immature DCs reside in the tissue where they collect antigenic material and are exposed to signals. Based on the combinations of signals received, maturation of the DCs occurs generating two terminal differentiation states, mature or semi-mature. Mature DCs have an activating effect while semi-mature DCs have a suppressive effect. The different output signals (termed output cytokines) generated by the two terminal states of DCs differ sufficiently to provide two different contexts for antigen presentation, shown abstractly in Figure 1.

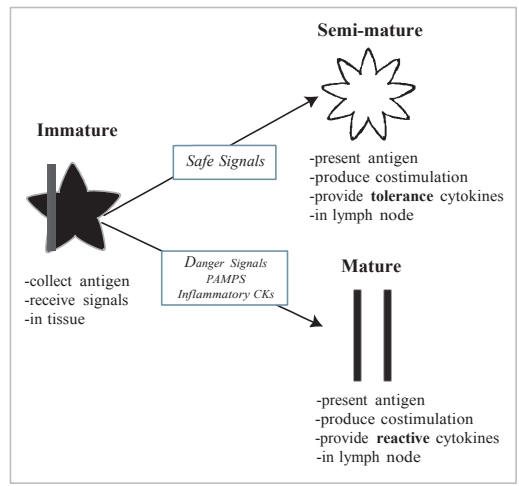

Fig. 1. An abstract view of DC maturation and signals required for differentiation. CKs denote cytokines.

The characteristics of the relevant signals are summarised below:

- PAMPS are pre-defined bacterial signatures, causing the maturation of immature DCs to mature DCs through expression of 'mature cytokines'.
- Danger signals are released as a result of damage to tissue cells, also increasing mature DC cytokines, and have a lower potency than PAMPs.
- Safe signals are released as a result of regulated cell death and cause an increase in semi-mature DC cytokines, and reduce the output of mature DC cytokines
- Inflammatory cytokines are derived from general tissue distress and amplify the effects of the other three signals but are not sufficient to cause any effect on immature DCs when used in isolation.

## 3 Dendritic Cells in silico

The Dendritic Cell Algorithm (DCA) was developed as part of the Danger Project[1], which aims to find the missing link between AIS and Intrusion Detection through the application of the danger theory[8]. The danger theory proposes that the immune system responds when damage to the host is detected, rather than discriminating between self and non-self proteins. The project encompasses artificial tissue[3] and T-cells[7], and the libtissue framework[11]. The DCs are the detection component developed within this project.

### 3.1 Libtissue

Libtissue is a software system which allows the implementation and testing of AIS algorithms on real-world problems based on principles of innate immunology [10], [11]. It allows researchers to implement AIS algorithms as a collection of cells, antigen and signals interacting within a tissue compartment. The implementation has a client/server architecture, separating data collection from data processing. Input data to the tissue compartment is generated by sensors monitoring environmental, behavioural or context data through the libtissue client, transforming this data into antigen and signals. AIS algorithms can be implemented within the libtissue server, as libtissue provides a convenient programming environment. Both client and server APIs allow new antigen and signal sources to be added to libtissue servers, and the testing of the same algorithm with a number of different data sources. Input data from the tissue client is represented in a tissue compartment contained on the tissue server. A tissue compartment is a space in which cells, signals and antigen interact. Each tissue compartment has a fixed-size antigen store where antigen provided by libtissue clients is placed. The tissue compartment also stores levels of signals, set either by tissue clients or cells.

### 3.2 Abstract View of the DCA

The DCA is implemented as a libtissue tissue server. Input signals are combined with a second source of data, such as a data item ID, or program ID number. This is achieved through using a population of artificial DCs to perform aggregate sampling and data processing. Using multiple DCs means that multiple data items in the form of antigen are sampled multiple times. If a single DC presents incorrect information, it becomes inconsequential provided that the majority of DCs derive the correct context. The sampling of data is combined with context information received during the antigen collection process. Different combinations of input signals result in two different antigen contexts. Semi-mature antigen context implies antigen data was collected under normal conditions, whereas a mature antigen context signifies a potentially anomalous data item. The nature of the response is determined by measuring the number of DCs that are fully mature, represented by a value, MCAV - the mature context antigen value. If the DCA functions as intended, the closer this value is to 1, the greater the probability that the antigen is anomalous. The MCAV value is used to assess the degree of anomaly of a given antigen. By applying thresholds at various levels, analysis can be performed to assess the anomaly detection capabilities of the algorithm.

The DCA has three stages: initialisation, update and aggregation. Initialisation involves setting various parameters and is followed by the update stage. The update stage can be decomposed into tissue update and cell cycle. Both the tissue update and cell cycle form the libtissue tissue server. Signal data is fed from the data-source to the tissue server through the tissue client.

The tissue update is a continuous process, whereby the values of the tissue data structures are refreshed. This occurs on an event-driven basis, with values for signals and antigen updated each time new data appears in the system. Antigen data enters tissue update in the same, event driven manner. The updated signals provide the input signals for the population of DCs.

The cell cycle is a discrete process occurring at a user defined rate. In this paper, 1 cell cycle is performed per second. Signal and antigen from the tissue data structures are accessed by the DCs during the cell cycle. This includes an update of every DC in the system with new signal values and antigen. The cell cycle and update of tissue continues until a stopping criteria is reached, usually until all antigen data is processed. Finally, the aggregation stage is initiated, where all collected antigen are subsequently analysed and the MCAV per antigen derived.

### 3.3 Parameters and Structures

The algorithm is described using the following terms.

- Indices:
  $i = 0, ..., I$ input signal index;
  $j = 0, ..., J$ input signal category index;
  $k = 0, ..., K$ tissue antigen index;
  $l = 0, ..., L$ DC cycle index;
  $m = 0, ..., M$ DC index;
  $n = 0, ..., N$ DC antigen index;
  $p = 0, ..., P$ DC output signal index.

- Parameters:
  I = maximum number of input signals per category;
  J = maximum number of categories of input signal;
  K = maximum number of antigen in tissue antigen vector;
  L = maximum number of DC cycles;
  M = maximum number of DCs in population;
  N = maximum number of antigen contained per DC;
  P = maximum number of output signals per DC;
  Q = number of antigens sampled per DC for one cycle.

- Data Structures:
  $DC_m = \{s^{DC}(m), a^{DC}(m), \bar{o}(m), t(m)\}$ - a DC within the population;
  $T = \{S, A\}$ - the tissue;
  S = tissue signal matrix;
  $s_{ij}$ = a signal type i, category j in the signal matrix S;
  A = tissue antigen vector;
  $a_k$ = antigen index k in the tissue antigen vector;
  $s^{DC}$ = DC signal matrix;
  $a^{DC}$ = DC antigen vector;
  o = temporary output signal vector for $DC_m$;
  o(m) = output signal p in the output signal vector of $DC_m$;
  $\bar{o}_p$ = cumulative output signal vector for $DC_m$;
  $t_m$ = migration threshold for $DC_m$;
  $w_{ijp}$ = transforming weight from $s_{ij}$ $o_p$.

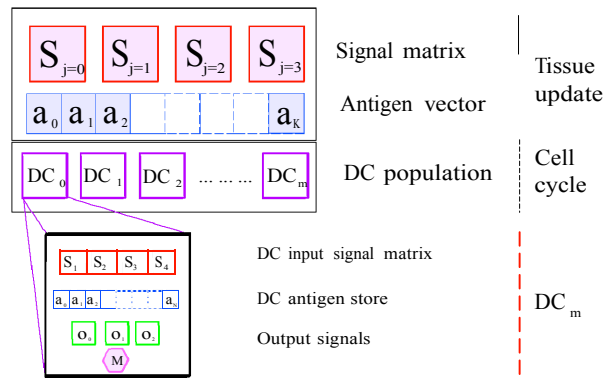

Fig. 2. Tissue and Cell Update components, where $S_{i,j}$ is reduced to $S_j$.

The data structures are represented graphically in Figure 2. Each $DC_m$ transforms each value of $s^{DC}(m)$ to $o_p(m)$ using the following equation with suggested values for weightings given in Table 1. Both the equation and weights are derived from observing experiments performed on natural DCs (personal communication from Dr J. McLeod and colleagues, UWE, UK), and information presented in Section 2 (more details found in [4]).

$$o_p(m) = \frac{\sum_i \sum_{j=3}^{j=3} W_{ijp} s_{ij}^{DC}}{\sum_i \sum_{j=3} |W_{ijp}|} * \frac{\sum_i W_{i3p}(s_{i3}^{DC}+1)}{\sum_i |W_{i3p}|} \quad \forall p$$

Table 1. Examples of weights used for signal processing

| $W_{ijp}$ | j = 1 | j = 2 | j = 3 | j = 4 |
|---|---|---|---|---|
| p = 1 | 2 | 1 | 2 | 1 |
| p = 2 | 0 | 0 | 3 | 1 |
| p = 3 | 2 | 1 | -3 | 1 |

The tissue has containers for signal and antigen values, namely S and A. In the current implementation of the DCA, there are 4 categories of signal (j = 3) and 1 signal per category (i = 0). The categories are derived from the 4 signal model of DC behaviour described in Section 2 where: $s_{0,0}$ = PAMP signals, $s_{0,1}$ = danger signals, $s_{0,2}$ = safe signals and $s_{0,3}$ = the inflammatory signal. An antigen store is constructed for use within the tissue cycle where all DCs in the population collect antigen, which is also introduced to the tissue in an event driven manner.

The cell cycle maintains all DC data structures. This includes the maintenance of a population of DCs, $DC_m$, which form a sampling set of size M. Each DC has an input signal matrix, antigen vector, output signals, and migration threshold. The internal values of $DC_m$ are updated, based on current data in the tissue signal matrix and antigen vector. The DC input signals, $s_{ij}^{DC}$, use the identical mapping for signal categories as tissue $s_{ij}$ and are updated every cell cycle iteration. Each $s_{ij}^{DC}$ for $DC_m$ is updated via an overwrite every cell cycle. These values are used to calculate output signal values, $o_p$, for $DC_m$, which are added cumulatively over a number of cell cycles to form $\bar{o}_p(m)$, where p = 0 is costimulatory value, p = 1 is the mature DC output signal, and p = 2 is the semi-mature DC output signal.

### 3.4 The DCA

The following pseudocode shows the initialisation stage, cycle stage, tissue update and cell cycle.

initialise parameters $\{I, J, K, L, M, N, O, P, Q\}$
while (l < L)
    update A and S

```
    for m = 0 to M
      for 0 to Q
        DC_m samples Q antigen from A
      for all i = 0 to I and all j = 0 to J
        s_{ij}^{DC} = s_{ij}
      for n = 0 to N
        DC_m processes a_{nm}^{DC}
      for p to P
        compute o_p
        ō_p(m) = ō_p(m) + o_p
      if o_0(m) > t_m
        DC_m removed from population
        DC_m migrate to Lymph node
  l++
```

analyse antigen and calculate MCAV

## 3.5 Lymph Node and Antigen Aggregation

Once $DC_m$ has been removed from the population, the contents of $a_n^{DC}$ and values $\bar{o}_{pm}$ are logged to a file for the aggregation stage. Once completed, $s_{ij}^{DC}$, $a_n^{DC}$ and $\bar{o}_{pm}$ are all reset, and $DC_m$ is returned to the sampling population. The re-cycling of DCs continues until the stopping condition is met ($l = L$). Once all data has been processed by the DCs, the output log of antigen-plus-context is analysed. The same antigen is presented multiple time with different context values. This information is recorded in a log file. The total fraction of mature DCs presenting said antigen (where $\bar{o}_1 > \bar{o}_2$) is divided by the total amount of times the antigen was presented namely $\bar{o}_1/(\bar{o}_1 + \bar{o}_2)$. This is used to calculate the mean mature context antigen value or MCAV.

## 3.6 Signals and Antigen

An integral part of DC function is the ability to combine multiple signals to influence the behaviour of the cells. The different input signals have different effects on cell behaviour as described in Section 2. The semantics of the different category of signal are derived from the study of the influence of the different signals on DCs in vitro. Definitions of the characteristics of each signal category are given below, with an example of an actual signal per category. This categorisation forms the signal selection schema.

- PAMP - $s_{i0}$ e.g. the number of error messages generated per second by a failed network connection
  1. a signature of abnormal behaviour e.g. an error message
  2. a high degree of confidence of abnormality associated with an increase in this signal strength
- Danger signal - $s_{i1}$ e.g. the number of transmitted network packets per second

1. measure of an attribute which significantly increases in response to abnormal behaviour
2. a moderate degree of confidence of abnormality with increased level of this signal, though at a low signal strength can represent normal behaviour.
- Safe signal - $s_{i2}$ E.g. the inverse rate of change of number of network packets per second. A high rate of change equals a low safe signal level and vice versa.
    1. a confident indicator of normal behaviour in a predictable manner or a measure of steady- behaviour
    2. measure of an attribute which increases signal concentration due to the lack of change in strength
- Inflammatory signal -$s_{i3}$ e.g. high system activity when no user present at a machine
    1. a signal which cannot cause maturation of a DC without the other signals present
    2. a general signal of system distress

Signals, though interesting, are inconsequential without antigen. To a DC, antigen is an element which is carried and presented to a T-cell, without regard for the structure of the antigen. Antigen is the data to be classified, and works well in the form of an identifier, be it an anomalous process ID[5] or the ID of a data item [4]. At this stage, minimal antigen processing is performed and the antigen presented is an identical copy of the antigen collected. Detection is performed through the correlation of antigen with signals.

## 4 Return of the Nmap - the Port Scan Experiment Revisited

The purpose of these experiments is as follows:

1. To validate the theoretical model which underpins the DCA
2. To investigate sensitivity to changes in the treatment of signals
3. To apply the DCA to anomaly detection for computer security

### 4.1 Port Scanning and Data

In this paper, port scanning is used as a model intrusion. While a port scan is not an intrusion per se, it is a 'hacker tool' used frequently during the information gathering stage of an intrusion. This can reveal the topology of a network, open ports and machine operating systems. The behaviour of outgoing port scans provide a small scale model of an automated attack. While examination of outgoing traffic will not reveal an intruder at the point of entry, it can be used to detect if a machine is subverted to send anomalous or virally infected packets. This is particularly relevant for the detection of scanning worms and botnets. The DCA is applied to the detection of an outgoing port scan to a single port across a range of IP addresses, based on the ICMP 'ping' protocol.

Data is compiled into 30 sessions, namely 10 attack, 10 normal and 10 control sessions. Each session includes a remote log-in to the monitored machine via SSH, and contains an event. The attack session includes a port scan performed by popular port scanning tool nmap, using the -sP option for an ICMP 'ping' scan, across a range of 1020 IP addresses. The normal session includes a transfer of a file of 2.5MB from the monitored machine to a remote server. The control session has no event and allows us to observe any signal deviations caused through monitoring the SSH session.

### 4.2 Signals and Antigen

Data from the monitored system are collected for the duration of a session. These values are transformed into signal values and written to a log file. Each signal value is a normalised real-number, based on a pre-defined maximum value. For this experiment the signals used are PAMPs, danger and safe signals. Inflammatory cytokines ($S_{i4}$) do not feature as they are not relevant for this particular problem. PAMPs are represented as the number of "destination unreachable" errors-per-second recorded on the ethernet card. When the port scan process scans multiple IP addresses indiscriminately, the number of these errors increases, and therefore is a positive sign of suspicious activity. Danger signals are represented as the number of outbound network packets per second. An increase in network traffic could imply anomalous behaviour. This alone would not be useful as legitimate behaviour can cause an increase in network packets. The safe signals in this experiment are the inverse rate of change of network packets per second. This is based on the assumption that if the rate of sending network packets is highly variable, the machine is behaving suspiciously. None of these signals are enough on their own to indicate an anomaly. In these experiments the signals are used to detect the port scan, and to not detect the normal file transfer.

During the session each process spawned from the monitored ssh session is logged through capturing all system calls made by the monitored processes using strace. Antigen is created with each system call made by a process, with antigen represented as the process ID value of a system call. Each antigen is processed subsequently by the DCA, and those presented with context are assigned a MCAV for assessment.

### 4.3 The Experiments

Experiments are performed to examine the influence of using different signal mappings. In these experiments a signal designed to be a PAMP is used as a danger signal and vice versa. The same is performed with PAMP and safe signals. We hypothesise based on previous experience using the DCA that it will be robust to incorrect signal mapping between danger and PAMP signals, but will lose detection accuracy if a safe signal is switched with a PAMP.

We also examine the effect of multiple antigen sampling on the performance of the algorithm. The DCA is designed so each DC can present multiple antigen on migration from the sampling population. Each DC presents a small subset of the total antigen within the tissue for its lifetime in the cell cycle. If multiple copies of the same antigen are used, robust coverage of input antigen can be achieved. To investigate the influence of multiple antigen presentation, an experiment is performed through limiting the antigen storage capacity (N) of each DC to 1. If less antigen is presented, the accuracy of the DCA could be impeded. An additional version of the DCA, known as 'DCLite', is implemented as the most basic form of the algorithm. DCLite uses one context signal, with N = 1, as in experiment M4. Based on our working knowledge of the data and of the DCA, we predict that it not possible to perform anomaly detection with the PAMP signal ($S_{0,1}$) alone. The performance of the algorithm under the various conditions is assessed through analysing the MCAV values. Five experiments are performed:

M1  using the suggested 'hand selected' input signals
M2  danger and PAMP signal swapped
M3  PAMP and safe signal swapped
M4  using a DC antigen vector size of 1, with signal mapping M1
M5  DC antigen vector of size 1 and using the PAMP signal only (DCLite)

Experiments M1 - M5 are performed for all individual attack and normal datasets as separate runs. Each data session is analysed by the DCA 3 times for each experiment (a total of 240 runs). Parameters for the experiments are as follows: $I = 1; J = 4; K = 500; L = 120; M = 100; N = 50; P = 3; Q = 1$. All experiments are performed on a AMD Athlon 1GHz Debian Linux machine (kernel 2.4.10) with all code implemented in C (gcc 4.0.2).

4.4  Results

The mean MCAV for each process type and each session type, both attack and normal, are recorded and presented in Table 2. Any process generating a non-zero MCAV is considered for analysis and termed a process of interest. The MCAV values for the 4 processes of interest for the attack sessions are represented in Figure 3. This shows experiment M1-M4 for the two normal processes of the bash shell (bash) and ssh demon (sshd) and the two anomalous processes namely the nmap and the pseudo-terminal slave (pts) which displays the nmap output. The MCAV values for the anomalous processes is significantly higher than that of the normal processes for experiments M1, M2 and M4. Experiment M3 does not show the same trend, though interestingly the nmap MCAV is not significantly different to the values for experiments M1, M2 and M4. All MCAV values for experiment M5 equal 1 because antigen is never presented in a semi-mature context due to lack of other signals. The normal session is represented in a similar manner, also shown in Figure 3. Significantly lower values for MCAV for all processes are reported, with the exception of experiment M3. The processes of interest include the bash shell, ssh demon, the file transfer (scp) and a forwarding client (x-forward). In the control experiment the mean MCAV values for all presented antigen were zero - no processes of interest could be highlighted. From this we can assume that the process of remote log-in is not enough to change the behaviour of the machine. All antigens were presented in a safe context implying steady-state system behaviour reflected through the MCAV output of the algorithm.

Table 2. MCAV values for each experiment across each dataset.

| Expt. | Attack | | | | | | | |
|---|---|---|---|---|---|---|---|---|
| | nmap | | pts | | bash | | sshd | |
| | mean | stdev | mean | stdev | mean | stdev | mean | stdev |
| M1 | 0.82 | 0.04 | 0.67 | 0.11 | 0.18 | 0.22 | 0.02 | 0.24 |
| M2 | 0.86 | 0.27 | 0.78 | 0.12 | 0.28 | 0.27 | 0.19 | 0.35 |
| M3 | 0.90 | 0.04 | 0.62 | 0.13 | 0.99 | 0.33 | 0.96 | 0.02 |
| M4 | 0.82 | 0.21 | 0.55 | 0.14 | 0.16 | 0.26 | 0.13 | 0.27 |
| M5 | 1.00 | 0.00 | 1.00 | 0.00 | 1.00 | 0.00 | 1.00 | 0.00 |
| Expt. | Normal | | | | | | | |
| | scp | | pts | | bash | | sshd | |
| | mean | stdev | mean | stdev | mean | stdev | mean | stdev |
| M1 | 0.14 | 0.29 | 0.12 | 0.25 | 0.01 | 0.02 | 0.01 | 0.01 |
| M2 | 0.24 | 0.33 | 0.18 | 0.29 | 0.04 | 0.03 | 0.05 | 0.09 |
| M3 | 1 | 0 | 1 | 0 | 1 | 0 | 1 | 0 |
| M4 | 0.19 | 0.25 | 0.1 | 0.17 | 0.01 | 0.03 | 0.05 | 0.08 |

4.5   Analysis

In experiment M1 distinct differences are shown in the behaviour of the algorithm for the attack and normal datasets. The MCAV for the the anomalous process is significantly larger than the MCAV of the normal processes. This is encouraging as it shows that the DCA can differentiate between two different types of process based on environmentally derived signals. In experiment M2 the PAMP and danger signals were switched. In comparison with the results presented for experiment M1, the MCAV for the anomalous process is not significantly different (paired t-test $p < 0.01$). However, in experiment M2, the standard deviations of the mean MCAVs are generally larger and is especially notable for the nmap process. Potentially, the two signals could be switched (through accidental means or incorrect signal selection) without altering the performance of the algorithm significantly. Experiment M3 involved reversing the mapping of safe and PAMP signals. The safe signal is generated continuously when the system is inactive and when mapped as a PAMP constantly generated full maturation in the artificial DCs, shown by the high MCAV value for all processes indiscriminately. Interestingly, in M3 the MCAV value for the anomalous processes in the attack datasets is lower than the normal process' value. For the normal dataset, all processes are classified as anomalous, all resulting in a MCAV of 1, a 100% false positive rate. These three experiments show that adding some expert knowledge is beneficial to the performance of the algorithm. It also supports the use of the proposed signal selection schema for use within the algorithm and has highlighted one key point - danger and PAMP signals should increase in response to a change in the system, whereas a PAMP must be the opposite, namely an indicator of little change within the system.

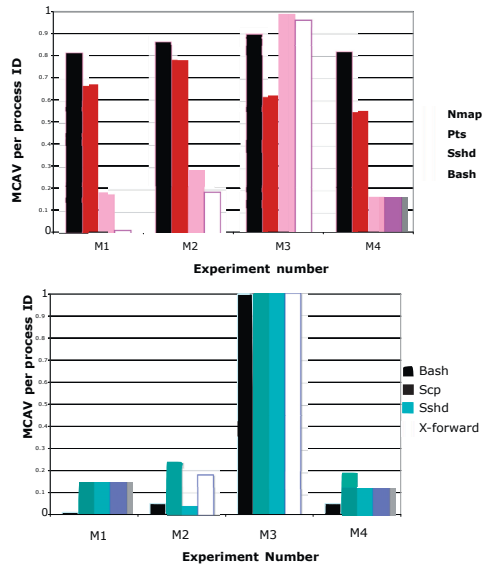

**Fig. 3.** The rate of detection for attack (upper graph) and normal (lower graph) for the 4 processes of interest (MCAV value) for experiments M1-M4 is shown.

By comparing the results from experiment M1 and M4, the influence of multiple antigen sampled per DC can be observed. In M4, the anomalous processes' MCAV are significantly greater than that of the normal processes. In comparison with M1, the detection of the anomalous processes was not significantly different for nmap, and was slightly lower for the pts process. Conversely, the MCAV for all normal processes from both the attack and normal datasets was greater than in experiment M1. Examination of the number of antigen presented revealed that fewer antigens per process were presented than in experiment M1. This implies that the MCAV values were generated from a smaller set size and could be responsible for the differences in detection. Multiple antigen sampling can improve the detection of anomalous processes while reducing the amount of normal processes presented as anomalous. More experiments must be performed using a range of antigen vector sizes to confirm this result. Experiment M5 yielded interesting results, showing it is not possible to discriminate between normal and anomalous (nmap) processes based on the PAMP signal alone. In M5, 3 out of the 10 datasets yielded no results, with insufficient PAMP signal generated to cause antigen presentation. For the remaining 7 datasets, all processes of interest produced a MCAV of 1. No discrimination was made between the normal and anomalous processes. In the absence of being able to discriminate based on the MCAVs, it may still be possible to determine the anomalous process for M5 based on the ratio of presented antigen to antigen input. The ratio for nmap antigen over the 7 successful runs is 0.054, and 0.02 for the ssh demon. A paired T-test shows that the sshd antigen ratio was significantly larger than the nmap ratio, further confirming the poor performance of DC Lite. One possible explanation for the poor performance of the DCA is that the safe signal is vital to provide some 'tolerance' for the processes which run constantly such as the ssh demon. Further investigations will be performed with the use of safe signals and the role of active suppression in the performance of the DCA.

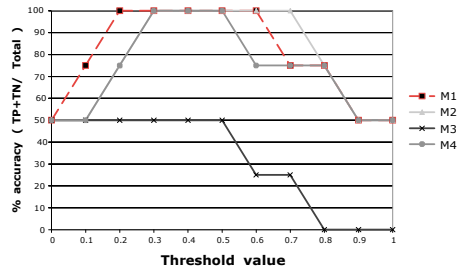

Fig. 4. Analysis of attack data for experiment M1-M4 in terms of accuracy at different thresholds

The accuracy for experiments M1-M4 is calculated by applying increasing threshold values to the MCAV values for the attack datasets, within a range of 0-1 at 0.1 intervals. If the MCAV value of a process exceeds this threshold then the process is classed as anomalous. The number of true positives and true negatives are calculated. The accuracy is calculated for each experiment (accuracy = true positives+true negatives / total number of processes) and the results of this analysis are presented in Figure 4. This figure shows that for experiment M1, if the threshold is between 0.2 and 0.7 the anomaly detection accuracy is 100%. For experiment M2 100% accuracy is also achieved, but is in the range of 0.3-0.8. M4 is of interest, as the range at which 100% accuracy is achieved is reduced in comparison to M1 and M2. As expected M3 performs significantly poorer than all others, also shown in Figure 4. For the normal dataset a similar analysis showed lower rates of false positives for increasing thresholds, with the exception of M3.

## 5 Conclusions

In this paper the DCA has been described in detail and interesting facets of the algorithm have been presented. The importance of careful signal selection has been highlighted through experiments. The DCA is somewhat robust to misrepresentation of the activating danger and PAMP signals, but care must be taken to select a suitable safe signal as an indicator of normality. In addition, the influence of multiple antigen presentation by each DC was investigated. Reduced antigen throughput, a decrease in detection of true positives and an increase in the rate of false positives are observed. The process by which these signals are combined has been described, and how changes in the semantic mappings of the signals influence the algorithm. Data processing was performed by a population of DCs, and multiplicity in sampling produced improved results. The baseline experiment highlighted that it is not possible to perform detection using a pre-defined 'signature-based' signal, regardless of how the results are analysed. Not only have we demonstrated the use of the DCA as an anomaly detector, but have also uncovered elements of behaviour previously unseen from the application of this algorithm.

Many aspects of this algorithm remain unexplored such as the sensitivity of the parameters and scalability in terms of number of cells and number of input signals. Our future work with this algorithm includes a sensitivity analysis and the generation of a solid baseline for comparison, in addition to performing similar signal experiments with a larger, more realistic, real-time problem.

## 6 Acknowledgements

This project is supported by the EPSRC (GR/S47809/01)